\documentclass[journal]{IEEEtran}
\usepackage[affil-it]{authblk}

\usepackage{blindtext}

\usepackage{url}
\usepackage{amsmath,amsfonts,amssymb}
\usepackage{mathptmx}
\usepackage{xcolor}
\usepackage{authblk}
\usepackage[latin1,utf8]{inputenc}
\usepackage[english]{babel}
\usepackage{lmodern}
\usepackage{siunitx}
\usepackage{textgreek}
\usepackage{gensymb}
\usepackage{textcomp}
\usepackage[version=4]{mhchem}
\usepackage[scaled]{helvet}
\usepackage[T1]{fontenc}
\usepackage{lettrine} 
\usepackage[rightcaption]{sidecap} 
\usepackage[misc]{ifsym} 
\usepackage{bbding} 
\usepackage{nameref}
\usepackage{cleveref}

\usepackage{svg}

\newcommand{\Real}[0]{\mathbb{R}}
\newcommand{\Natural}[0]{\mathbb{N}}

\begin{document}

\title{Deep Neural Patchworks: \\  Coping with Large Segmentation Tasks}

\author[1,2,3]{Marco Reisert}
\author[1,3]{Max Russe}
\author[1,4]{Samer Elsheikh}
\author[1,3]{Elias Kellner}
\author[5]{Henrik Skibbe}

\affil[1]{\small{Medical Faculty of Freiburg University, Medical Center – University of Freiburg, Germany}
}

\affil[2]{\small{Department of Stereotactic and Functional Neurosurgery}
}

\affil[3]{\small{Department of Diagnostic and Interventional Radiology}
}

\affil[4]{\small{Department of Neuroradiology}}

\affil[5]{\small{RIKEN Center for Brain Science, Brain Image Analysis Unit, Japan}}

\maketitle

\begin{abstract}
Convolutional neural networks are the way to solve arbitrary image segmentation
tasks. However, when images are large, memory demands often exceed the available resources, in particular on a common GPU. Especially in biomedical imaging, where 3D images are common, the problems are apparent. A typical approach to solve this limitation is to break the task into smaller subtasks by dividing images into smaller image patches. Another approach, if applicable, is to look at the 2D image sections separately, and to solve the problem in 2D. Often, the loss of global context makes such approaches less effective; important global information might not be present in the current image patch, or the selected 2D image section. 
Here, we propose Deep Neural Patchworks (DNP), a segmentation framework that is based on hierarchical and nested stacking of patch-based networks that solves the dilemma between global context and memory limitations. 
\end {abstract}



\section*{Introduction}
Attention plays an important role in how we process visual information. It is the process of selectively focusing on a specific aspect of information. Here, we present Deep Neural Patchworks (DNP), a novel network architecture that exploits the principles of hierarchical attention to seek for important information in images in a coarse-to-fine manner. It shares similarities with hierarchical approaches \cite{shen2017deep}, which popular \cite{zhang2020munet,gadermayr2019cnn,vu2019tunet} and has also relationships to masked R-CNN \cite{he2017mask} which is a popular approach for object detection.

If we think about how humans handle very specific visual recognition
tasks, it is quite nearby to understand the proposed principle. Think
of a radiologist trying to identify kidney stones in a whole body CT image. His focus of visual
attention traverses in a top-down fashion, first searching for the approximate location 
of the kidney, then, capturing the kidney, and finally looking at the typical locations
within the organ for the stones itself. All these steps require different levels of details. While at first a rough image of the entire organs is sufficient to spot the kidney, a much higher image resolution is required to locate the stones.

The DNP mimics this behavior. It first processes a large portion of the entire image at a very rough scale. Then it decides which part is worth further exploration. This process is repeated several times in a stochastic manner. The matrix sizes of the inputs (patches) are typically small (around $32^2$ pixels, or $32^3$ voxels, respectively), and remains constant among all scales. This means, while the field of view is wide at the upper levels, the spatial resolution is low. On the other hand, in the lower levels, the spatial resolution is getting higher, but due to the narrow field of view, the direct image information is not enough to resolve global context. DNP augments the input with information from the preceding levels.

A key feature of DNPs is that the exploration is typically stochastic, and DNP repeats the nested patching process several hundred times in order to obtain a result. The stochastic nature and the large amount of repetitions strengthens a DNP against outliers. Practically, the sampling of the patch locations can be random, or organized in a more tree-like fashion. By introducing slight distortions to the process of cropping patches out of the original image, geometric augmentations, like rotations, stretchings and flipping are inherently built in.


With patchworks, the complexity and the dimensionality of the networks are shifted. Instead of having a few large examples, data can be chunked into
much smaller pieces, but still retaining context and global information accessible. Matrix sizes of the patches can be chosen much smaller, while still keeping the influence of global information. 

\begin{figure*}
    \centering
    \includegraphics[width=1.8\columnwidth]{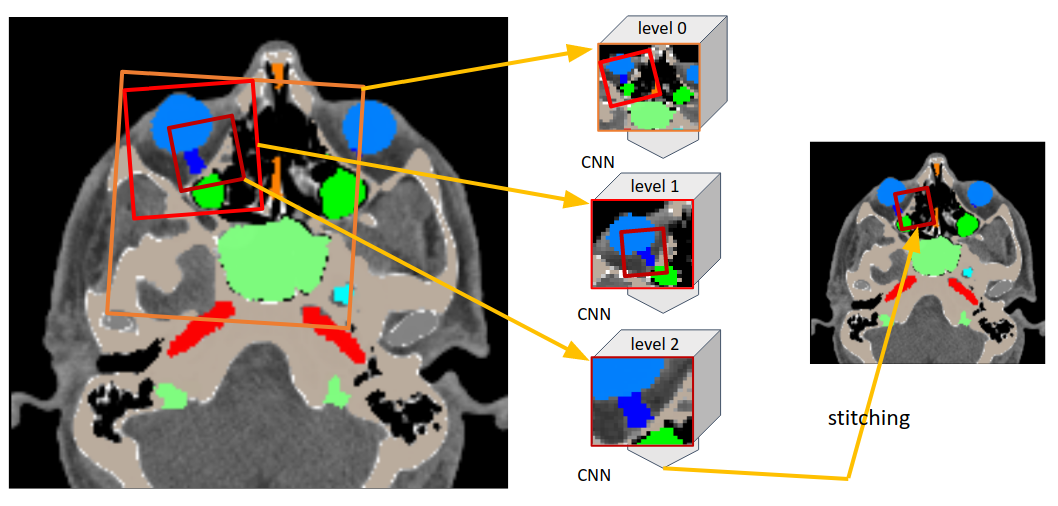}
    \caption{
The basic principle of patchwork: the framework uses nested patches of typically fixed matrix size, but decreasing physical size. In each scale, any type of architecture can be used (like a U-net, \cite{falk2019u}). Typically, matrix sizes are in the range of $32^2$ or $32^3$ voxels for all scales. In the above example, a scale pyramid of depth three is depicted. The scale of the intermediate levels are exponentially interpolated. The information flow is from the coarsest to the finest level. The initial large patches (maybe also the full image) contain coarse but global information, the small patches also integrate high resolution information and get the global information from the coarser scales. }
\end{figure*}

In this short report, we propose the patchwork toolbox implementing all the core features of the framework and comprises a python application interface. The toolbox is implemented in tensorflow and tailored to medical imaging semantic segmentation tasks. It understands the NIFTI format (based on nibabel).

\section*{Patchwork Architecture}
In the following, we consider the 2D scenario $d=2$ for convenience. 
A patch $p=(A,s)$ is characterized by a patch shape $s \in \Natural^d$, and a
homogeneous matrix $A\in \Real^{{(d+1)}\times{(d+1)}}$. Patches are parts of a larger image, and the matrix $A$ defines the location within that image. Without restriction of generality, we can characterize the entire input image as a patch as well.

The matrix $A$ maps the local voxel coordinate system of the patch 
to an absolute coordinate system (e.g. a physical coordinate in millimeter). 
The data corresponding to a patch is a $d+1$ dimensional array $D$ of 
shape $(s_0,s_1,f)$, where $s=(s_0,s_1)$ are the spatial dimensions (the shape) and $f$ is the feature dimension.

patch shape $s \in \Natural^d$. The input image is characterized by 
the same properties and is in the following treated as a patch without 
restriction of generality. The matrix $A$ maps the local voxel coordinate system of the patch 
to an absolute coordinate system (e.g. a physical coordinate in millimeter). 
The data corresponding to a patch is a $d+1$ dimensional array $D$ of 
shape $(s_0,s_1,f)$, where $s=(s_0,s_1)$ are the spatial dimensions (the shape) and $f$ is the feature dimension.

\newcommand{\Rg}{\mathcal{R}}
\newcommand{\B}{\mathcal{B}}
\newcommand{\Z}{\mathcal{Z}}

A re-sampling operation $\Rg$ takes patch properties $p$ and data $D$ of an existing
patch and evaluates the data at new locations given by other patch properties $p'$. 
Formally, the re-sampling can be written
\[
D'[v,:] = D[A_{p'}^{-1} A_p v,:]
\]
where $v$ denotes the homogeneous voxel coordinate. 
The re-sampling step is either involved during a crop of a subpatch out of
an existing patch, or during scattering patch information back into the 
full image. The practical implementation of both, differ, but for shorthand, 
we do not differentiate here and just write $D' = \Rg((p,D),p')$. 

The hierarchical architecture of the network can be
conveniently written in a recursive way from coarse to fine. Assume a given 
input image $(p_0,I_0)$ and a series of patching properties $p_1,\hdots,p_{N-1}$, then the patchwork does the following 
\begin{eqnarray}
      I_n &:=& \Rg((p_0,I_0),p_n) \\
      C_n &:=& \Rg((p_{n-1},X_{n-1}),p_n) \\ 
      X_n &:=& \B_n([I_n,C_n]) 
\end{eqnarray}
where $I_n$ is the input data of the current layer, $C_n$ the crop of the 
forwarded output from the last layer (initially $C_0=[]$) and $X_n$ the output of the network 
in the current layer. The operator $\B_n$ can be any 
kind of (convolutional) network that maps images to images. The $[\cdot,\cdot]$ operation is a simple concatenation along the feature dimension. 
The network's final output is $X_{N-1}$, but also intermediate outputs $X_n$ with $n<N$ can be used for loss computations. 
The simple concatenation $[I,C]$ of forwarded outputs and input is just one option, 
there are a multitude of possibilities here.

In conclusion: given an input image $I_0$ and patching properties,
$P=(p_1,\hdots,p_{N-1})$ we define the network response to be 
$\Z(I_0,P) := X_{N-1}$. 
As mentioned, the network processes numerous image patches, and many of them overlap. For finalizing the output, the results have to be merged, and rendered 
back into a full image. Formally we can write this scatter operation as
\[
  Y = \frac{\sum_{(p_1,\hdots,p_N) = P \in \mathcal{P} }  \Rg(\Z(I_0,P),p_N) }
                {\sum_{(p_1,\hdots,p_N) = P \in \mathcal{P} }  \Rg(1,p_N) }
\]
where we divide by the number of outputs a voxel received.

\begin{figure*}
    \centering
    \includegraphics[width=1.8\columnwidth]{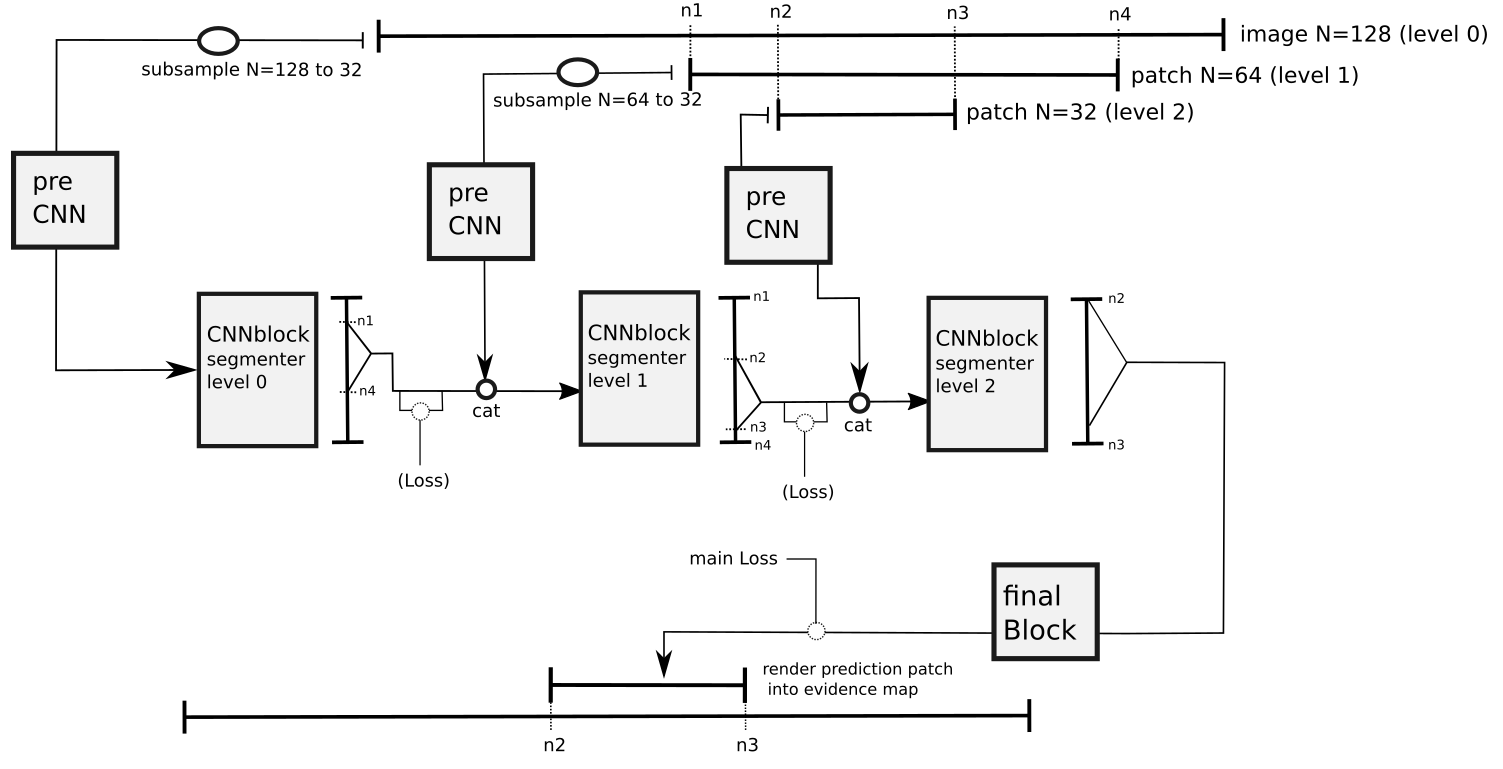}
    \caption{A sketch of the working principle of patchwork in 1D. The idea is close to other hierarchical approaches, but only sub patches are forwarded for further processing to the next level. The sub patches are contained in its parent patch and the information, which is in spatial correspondence to its children, is forwarded from the parent to its children. The loss can be taken at intermediate levels, or only after the final stage. }
\end{figure*}

\subsection*{Cropping and Stitching}

Cropping denotes the operation of extracting patches from the image. With stitching, we place patches back into an image. It should be noted that in both operations, patches may overlap thus during stitching, multiple patches can contribute to the same image content. 

The crop of a patch is implemented by a gathering operation, where index arrays are used for lookup (tensorflow's \verb+gather_nd+). On the other hand, the composition of 
output patches is a scatter operation  (tensorflow's \verb+scatter_nd+), where the index arrays determine
the target position in the output image.
The cropping operation can be implemented by simple nearest-neighbor interpolation or by a (tri-)bilinear interpolation scheme (\verb+interp_type+). Also, a proper smoothing of the input before the actual cropping might be an option( \verb+smoothfac_data+)

In a simple experiment, we found that nearest-neighbor interpolation is superior in terms of reconstruction quality. When setting $\B_n$ to the identity, i.e. $\B_n([I,D]) = I$, the stitched image $Y$ is just a downgraded original image. When using NN-interpolation, the reconstructed image is typically closer to the original image than for linear interpolation.

\subsection*{Generation of patches}

Two patching schemes (\verb+generate_type+) are implemented: a random distribution of patch centers and a tree-like sampling. During random sampling, patch centers are uniformly distributed 
within its parent patch. If the child patch is not fully contained within its parent, the center is adjusted such that the borders of the child
snap onto the border of the parent (\verb+snapper+). This ensures that the boundary regions are sampled dense enough, which avoids boundary artifacts. Faced with an unbalanced problem, label arrays can be used as spatial probabilities for the patch center distribution to represent regions of interest more intense  (\verb+balance+)

Tree-like sampling ensures a full coverage of the 
considered image, however introduces also a certain locational bias. To mitigate the bias, patch centers can be additionally distorted from its ideal position by noise and sampled multiple times.

Besides the patch shapes and the depth $N$, the patch sizes of a patching scheme $P$ is one major hyperparameter of the architecture (\verb+scheme+). It is quite nearby to use fixed scaling factors along
each dimension. Then, the initial patch size and the output pixel/voxel size determine all intermediate patch sizes. The initial patch size can be given in the absolute physical system, or just relative to the FOV of the data. In the same way, output pixel/voxelsize might be chosen absolute or relative to the input sizes.

Note that drawing patches from the original data and the preceding training procedure is naturally parallelizable. In present implementation it is possible you can fork a child process which is responsible for the cropping part. The child process is soley using the CPU and not using the GPU.

\subsection*{Augmentation}

Due to the architecture of the patching approach, spatial augmentations are integrated seamlessly (\verb+augment+). Orientation and scalings are naturally represented
by the homogeneous representation $A$ of the patch: we parametrized the affine part $A[1:d,1:d]$ as
\[
A[1:d,1:d] = R_1 \Sigma R_2
\]
where $R_1$ and $R_2$ are rotations and $\Sigma$ a diagonal matrix. In 3D the rotations are generated by normally
distributed quaternions, in 2D just by a normally distributed angle. Scalings and flips are simple manipulations of the diagonal of $\Sigma$. 

Note that we have the ability to use different transformations at different levels of the patchwork (\verb+independent_augmentation+). Depending on the level, augmentation strengths may also differ. Coarser levels might experience fewer rotations than finer levels. In fact, this approach can mimic deformable transformations and allows controlling of local versus global invariance. Locally, the problem could be orientationally invariant (e.g. the detection of vessels), but 
globally one might be only interested in vessels in
a certain body part, which has always upright 
orientation.

\subsection*{Training}
The input of one training sample is formed by a stack of  patches, where each patch corresponds to an input for one layer, and target patches. There are two options for the target patches. Either a stack of target patches (one per layer), or only a single target patch for the output of the last layer. Correspondingly, the loss is either taken at each layer (e.g. for  the leading feature dimensions) or simply at the final level (\verb+intermediate_loss+). 
The CNNs in each layer can be independent (also different architectures) or they can all be identical and share their weights (\verb+identical_blocks+). 

The basic training consists of two nested loops. Within the outer loop, patches are randomly drawn from the set of training images and form a patch set. The inner loop performs gradient descent on this patch set. Depending on the size of the training set (number of images), the number of patches per image can be varied to reach suitable sized patch sets, which fit into GPU memory. Depending on the size of the patch set one more epochs can be used for training. As a rule of thumb, time spent on patching and training should be roughly equal. 

There are different options for balancing the drawn patches. As already described above,
we can use the labels as spatial probability distributions to draw patch centers from (\verb+balance+). If we think of a task consisting of two labels, one very large label (e.g. an organ) and a small lesion which is not directly associated with the organ, we have to bit more careful. To balance between the large and small label, and get approximately the same number of 
samples, it is resonable to draw samples with a probability inversely proportional to the volume of the samples (\verb+autoweight+).

If there are very rare samples, a hard mining technique 
\verb+hard_mining+ is implemented on the patch level. That is, after GPU training, the hard samples (in terms of loss or DICE score) are kept for the next round and joined with some newly drawn patches. This is also useful when there are very rare labels. Imagine you have a set 100 training images and only 5 of them contain a certain label (like a cyst), then patchwork can automatically retain these rare examples and accumulate patches, which contain the cysts. 

In case of many labels (>50) even the proposed architecture can get into memory trouble. The final prediction maps can become enormous in size, e.g. think of large CT with 50 feature channels, one can easily create buffers large 32G for a single prediction. Therefore, patchwork has implemented a candidate sampling approximation \cite{candidatesampling} of the usual categorical cross entropy loss. The labels/classes are embedded in a low-dimensional feature space, which is small enough that the full prediction image is of reasonable size. In a final step, classes are assigned by searching for maximal responses.

\subsection*{Application}

\begin{figure*}
    \centering
    \includegraphics[width=1.8\columnwidth]{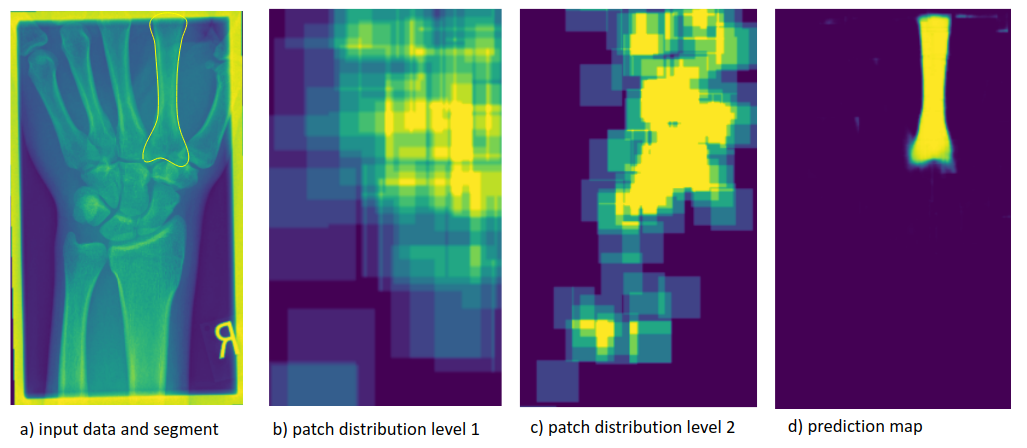}
    \caption{Patch Distributions for a 3-level patchwork architecture in 2D. In a) the x-ray projection of a right hand is shown together with the finger which we aim to segment. The size of the first patch is 70\% of the field of view. In b,c) we show how the 'lazy evaluation feature' distributes patches only in the areas, which are likely to contain the target structure (the distribution in level 0 is fully random and not shown). In d) we show the predictions in level 2.} \label{fig:hand}
\end{figure*}

Like in training, we can use a random or a tree-like 
distribution of patches. 
However, for small patches in the finest layers, full tree-like sampling schemes can get quite costly. With random sampling, it is 
difficult to guarantee full coverage of the image.
Fortunately, it is not necessary for most tasks to distribute patches exhaustively over the whole image (\verb+lazyEval+). If we train with intermediate losses, the intermediate outputs $X_i$ at coarse levels can already give evidence for the occurrence of the target within the patch. To measure evidence, we can simply
reduce the corresponding feature dimension by a sum or a maximum. 
Keeping only those patches with the highest evidence
values allows the patchwork to concentrate on the relevant
regions. In practice, we start to distribute $n$ patches 
uniformly at the coarsest level. Then, we keep the most
promising $n\alpha$ patches, where $0<\beta<1$ is a concentration
parameter (typically $\beta=0.25$ or $\beta=0.5$).
To keep the number of remaining patches the same the promising ones have $1/\beta$ children (\verb+branch_factor+). For tree-like sampling $1/\beta$ is the branching factor of the tree. This principle is
simply applied at each level. 
In Figure \ref{fig:hand} we show an example in 2D and how the patches are distributed (for \verb+branch_factor+=2 and $\beta=0.5$).
Of course, there are tasks where
this principle can fail. For example, if we search for two(!) structures, but one of the structures is not well discernible,
the network focuses on the good one, and the bad one is undersampled.

To avoid spurious responses caused by voxel that received only a few outputs from single patches, we modified the mean computation. Instead of dividing by the number $n$ of received outputs, we divide by $\sqrt{n^2+3\alpha}$. For $\alpha=1$ it ensures that voxels which received just a single output will never give a response $>0.5$.

\section*{Discussion}

Patchwork is an abstraction of most of the common types of architectures used in  biomedical image segmentation. It includes ordinary patch-based approaches, hierarchical approaches, and a continuous transition from 
full 3D approaches to 2D. Thereby, it solves
the dilemma of global awareness versus receptive field size, and hence, the typical memory problems are not so apparent anymore. Images of arbitrary size can be chunked such that
global context and fine details are preserved. Geometric augmentations come for free and are just part of the principle. 

But there is no free lunch: the inductive bias introduced by the 
approach is obvious. Due to its top-down
architecture, fine texture details do not have an impact on global decisions. For example, if you want to segment cells of type A and type B, but A and B differ only by a subtle, high resolution feature, patchwork will have severe problems, because the coarse levels
are not able to 'see' the subtle features. Patchwork uses high frequency details mostly for refinement. However, for most real world tasks, the bias works for us.

We experienced that decision values of patchworks are not so decisive compared to 
ordinary approaches. Usual approaches
tend to produce relatively hard decisions, and the transition from $p=0$ to $p=1$ is quite narrow. Patchwork shows a different behaviour and gives smoother maps. A reason for this 
 is the probabilistic nature of the approach. Even without any augmentation, a voxel does not receive one output, but a bunch of outputs stemming from different receptive patches, which are averaged.

Patchworks solve the problem that ordinary U-Nets have at the boundary of the domain, which are typically susceptible to artifacts. For patchworks every voxel/pixel might be a boundary voxel and the network trains avoiding these artifacts. This idea breaks partially translation covariance, but the random sampling works against any spatial priors.

We used in each level of the patchwork
an independent network $\B_i$. It is quite nearby to let these networks share their weights, i.e.
$\B_n=\B_k$, then, the patchwork architecture tends to become scale invariant. In fact, the number of level do not even have to be a fixed property of the network anymore.

\section*{Application Interface}
The \verb+tensorflow+ implementation of the patchwork toolbox can be found for download here \url{https://bitbucket.org/reisert/patchwork/wiki/Home}. There are also several examples as jupyter notebooks available. An alternative \verb+pytorch+ implementation can be found here: \url{https://bitbucket.org/skibbe/deep_patchwork}.

In the following, we shortly describe its application interface of the \verb+tensorflow+ implementation. 
The python API to patchwork is built on mainly two classes, one class for the implementation of the cropping procedure (\verb+CropGenerator+) and one class for the network itself (\verb+PatchWorkModel+). The class for cropping is mostly independent of the model class, which makes it also usuable in different contexts. In the following, we explain the most important parameters of the two classes. 

\subsection{Cropping the patches}

Here we report all options to the \verb+CropGenerator+ class defining the patching scheme
\begin{itemize}
\item \verb+depth+, the number of hierarchical levels, an integer larger than zero.
\item \verb+ndim+, the dimension of the network, 2 or 3

\item \verb+scheme+, a dictionary defining the patching scheme. It has to contain the 
\verb+patch_size+, a tuple defining the matrix size of the patches. It can also be a list of tuples,
if the matrix sizes vary over the levels. It may contain \verb+out_patch_size+, if the network changes the matrix size. The dictionary has to contain \verb+fov_mm+ or \verb+fov_rel+. Both tuples determine the size of the most outer patch. In case of \verb+fov_mm+ is passed, the size is given absolute in millimeter. In case of \verb+fov_mm+, the size is given relative to full size of the image (as ratios between 0 and 1). The same principle applies to \verb+destvox_mm+ and \verb+destvox_rel+, respectively, which determine the physical voxel size of the final patch in either millimeter or relative to the voxel sizes of the input images.
\item \verb+system+, a string, which is either 'world' or 'matrix'. Is the (non-augmented) patch aligned with the coordinate system defined by the imaging matrix ('matrix') or by physical coordinate system ('world')
\item \verb+snapper+, a list of length \verb+depth+ defining whether the patches are strictly
inside the parent patch (\verb+snapper+[i] = 1) or that only the cnter of the patch is inside parent patch.
\item \verb+interp_type+, a string which is either 'NN' for nearest neighbor interpolation or 'lin' for bi/tri-linear interpolation. It refers to the type of interpolation used during gathering the data (either for the original image or from the parent patch).
\item \verb+scatter_type+, a string which is either 'NN' for nearest neighbor interpolation or 'lin' for bi/tri-linear interpolation. It refers to the type of scattering operation when the results/network outputs are scattered back into the full image.
\item \verb+categorial_label+,
a list defining the label index mapping, which assign to the integer label in the label array a internal label index. Suppose, you have 'atlas' nifti, and you want to learn the anatomical labels 4, 16, and 30, just pass [4,16,30] and the network will be trained on this 3-class problem. If this parameter is None, it is assumed that multi class labels are encoded by the 4th dimension, i.e. training a n-class problem means shape[-1] = n for the shape of the label array.
\item \verb+categorical+,
a boolean indicating whether the multi labels are mutual exclusive (in case of categorial crossentropy/softmax like problems)
\item \verb+num_labels+,
the number of labels. Typically, the size of the 4th dimension (for non-categorial) or 
len(\verb+categorial_label+). Automatically set, if a model instance is present.
\item \verb+smoothfac_data+,
a float or a string. If a float is given, it refers to the width of the Gaussian used for smoothing prior to cropping the patches from the data. The width is relative the undersampling of the crop.
If \verb+smoothfac_data+=0 no smoothing is performed.
Other options are 'boxcar', convolution with a boxcar, 'max' max-pooling, 'mixture' a concatenation of boxcar, max-pooling and min-pooling.
\item \verb+smoothfac_label+,
same like \verb+smoothfac_data+ but for the labels.
\item \verb+normalize_input+,
None or a string. Either the whole image is multiplicatively normalized by the maximum ('\verb+max+') or the mean ('\verb+mean+'). Subtraction of the full image mean and division by standard deviation ('\verb+m0s1+') or patch-wise ('\verb+patch_m0s1+').
\end{itemize}

\subsection{The Network}

The \verb+PatchWorkModel+ class is derived from tensorflow model and contains the network training and application/deployment parts. In the following we report the most important parameters to the constructor of the class. 

\begin{itemize}
\item \verb+blockCreator+, a function returning a keras layer and taking as input the patching level, the output feature dimension the CNN should have a that level and the input shape of the network.
\item \verb+preprocCreator+,
preprocCreator=None,

\item \verb+forward_type+, a string, either 'simple', or 'bridge', or 'mult'
\item \verb+identical_blocks+, a boolean, if True always the same CNN block is used in all levels.
\item \verb+intermediate_out+, the number of output channels for intermediate levels is 
\verb+intermediate_out+ + \verb+num_labels+
\item \verb+intermediate_loss+, a boolean, whether the loss is applied at all levels
\item \verb+block_out+, either None, or a list of integers, which overwrite the number output channels in each level (given by \verb+intermediate_out+ + \verb+num_labels+).
\item \verb+finalBlock+, a keras layer, which follows the network (typically only appled after the finest level)
\verb+finalizeOnApply+, a boolean, whether the \item \verb+finalBlock+ is only applied during training.
\verb+finalBlock_all_levels+, a boolean, whether the \item \verb+finalBlock+ is applied on
all level, or just the final level.
\verb+crop_fdim+, None or an integer list, if set it crops the list of specified indices out of the 4th dimension of the data.
\verb+num_labels+, the number of labels. Typically, the size of the 4th dimension (for non-categorial) or len(\verb+categorial_label+).
\end{itemize}

\subsection{Training}

Here we report the most important parameters of the training procedure. The training consists of two nested loops. Within the outer loop, patches are randomly drawn from the set of training images. 
The inner loop performs gradient descent on the drawn patches.
\begin{itemize}
\item \verb+num_its+, number of iterations of the outer loop
\item \verb+epochs+, number of iterations of the inner loop
\item \verb+num_patches+, the number of patches drawn from each image
\item \verb+batch_size+, integer, the batch size
\item \verb+balance+, a dictionary controlling how patches are drawn from image and subpatches. Typically, the labels are used to draw more often from area with labels. 
The entry \verb+ratio+ (a float) controls how often the center of the patch is drawn from on-label voxel. For multi-label problems the \verb+label_weight+ parameter determines the weighting of the different labels (a list of floats). To automatically choose the weights such that the occurrence of different labels is approximately equalized pass \verb+autoweight:1+.
\item \verb+hard_mining+, in hard-mining mode patches from the last outer loop iteration are partially kept depending on their loss/f1/balance value. If \verb+hard_mining+>0 it determines the ratio of how many patches ar kept for the next training round.
\item \verb+hard_mining_order+, a string, which determines the criteria what type of patches are kept. Options are 'loss': the samples with the highest loss value are kept, 'f1': the samples with lowest 'f1' scores are kept, and 'balance': the rare samples are kept (in terms of label occurrence).
\item \verb+hard_mining_maxage+, integer, samples are kept as most as \verb+hard_mining_maxage+ rounds.

\item \verb+augment+, None or dictionary containing information for augmentation. For rotation augmentation set \verb+dphi+ to be a float (the angle width of the distribution in radians), or optionally in 3D a 3-tuple determining the angles corresponding to the rotation axis (actually the quaternion). A 2-tuple (2D) or 3-tuple (3D) for \verb+flip+ containing zeros or ones to enable flip augumentation in the corresponding dimension. A 2-tuple (2D) or 3-tuple (3D) for \verb+dscale+ containing float indicating an expected ratio for scaling augmentation. If \verb+independent_augmentation+ is set to False, all patch levels share the same augmentataion/transformation parameters, otherwise they are independently drawn. 

\item \verb+dontcare+, if set to True, for NaN/-1 values (float/integer) no loss is computed.

\item \verb+loss+, None or an array of functions of length \verb+depth+ returning for each level the loss function. 

\item \verb+optimizer+, None or the keras optimizer you want to use.

\item \verb+parallel+, a boolean indicating whether cropping of patches and training is running concurrently. Patchwork uses the \verb+forkserver+ start method for launching a process which does the cropping work. Note that the 
entrypoint script has to be appropriately prepared to avoid running the main process code when called from the child process.


\end{itemize}
\subsection{Application}

\begin{itemize}
\item \verb+generate_type+, there are two ways to distribute the patches spatially, either randomly \verb+random+ (but driven by the outputs of previous level), or in a systematic n-tree like fashion (\verb+tree+). 
\item \verb+num_patches+, in case of \item \verb+generate_type='random'+ this parameter controls how many patches are drawn in the first level.
\item \verb+branch_factor+, an integer, in case of \item \verb+generate_type='random'+ this is telling how many children a patch has (default is 1). 

\item \verb+num_chunks+, an integer determining the number how often the whole process is repeated. Used to get robust results and full coverage.

\item \verb+scale_to_original+, a boolean, whether the affine geometry is used, or the actual output dimension of the network.
\item \verb+augment+, the dictionary defining the augmentation behavior during application of the network. The same structure as in \verb+CropGenerator+. By default, it assumes the setting used during training. To turn off augmentation, just pass an empty dictionary.
\item \verb+ce_threshold+, this minimal probability to accept a decision different from background (in case of categorical loss types).

\item \verb+lazyEval+, a dictionary controlling how patches at lower levels are discarded, i.e. do not have any further children. 
The entry \verb+fraction+ tells how many patches are kept at each level.
Which patches are kept is controlled by an attention value computed from the output of the block at the current level. The patches with the highst attention values are kept. The attribute \verb+reduceFun+ controls how the output of the network is reduced to obtain the attention value (default is \verb+tf.reduce_mean+). The \verb+attentionFun+ is a function applied on the output (logits) of the current block (by default a sigmoid) prior to reduction. 
For example,  \verb+lazyEval={'ratio':0.5}+ and 
\verb+branch_factor=2+, keeps the number of patches evaluated in each level the same.

\item \verb+window+, a string telling how the patch window is spatially weighted during stitching the final prediction/output. Options are \verb+cos+ for cosine weighting and \verb+cos2+ for squared cosine weighting.
\item \verb+sparse_suppression+, a float, the parameter $\alpha$ avoiding spurious responses (default is 0).
\item \verb+out_typ+, a string defining how the results are written to the final nifti. Options are \verb+int16+,\verb+uint8+,\verb+float32+, 
which output the probabilities in different data types and different classes are encoded in the fourth dimension. The option \verb+mask+ results in thresholded maps at the optimal threshold on the validation set, or, if no validation set was available at 0.5. You can also set the thresholds manually by writing \verb+mask:0.4,0.5+, for example, in case of two classes. If you want to save in index notation without overlapping classes use \verb+atls+. With the same notation (e.g. \verb+atls:0.4,0.5+) you can decide for thresholds.
\item \verb+label_names+, a list of strings, if you have string labels for your classes pass them here, they get into the extended header as XML code readable from e.g. the HCP workbench or NORA (www.nora-imaging.org).

\item \verb+sampling_factor+, a float which is default 1. With this parameter, you can change the target voxel size. The new target voxel size is \verb+vsz/sampling_factor+. Note that all other parameters stay the same, only the final prediction volume where all patches are scattered into changes its dimension. You can use this to avoid stripe artifacts for rotated pattern.

\item \verb+level+, an integer which is default -1, or the string \verb+mix+. For debugging you can change here which level is finally scattered into the prediction volume (-1 for last level). If  \verb+level='mix'+ all levels are mixed, i.e. in regions that are not covered in the final level, lower levels are used by using an interpolation scheme. Note that you have to train with \verb+intermediate_loss+.

\end{itemize}

\bigskip
\noindent\textbf{Acknowledgements}\newline
We thank all the medical doctors (and applicants) that continuously use patchwork at the university medical center. With comments and bug reports, they help patchwork to become a valuable tool for medical image segmentation. This work was supported by the program for Brain Mapping by Integrated Neurotechnologies for Disease Studies (Brain/MINDS) from the Japan Agency for Medical Research and Development AMED (JP15dm0207001).

\bibliographystyle{plain}






\end{document}